\pdfoutput=1

\documentclass[11pt]{article}

\usepackage{acl}
\usepackage{clrscode3e}
\usepackage{times}
\usepackage{latexsym}
\usepackage{amssymb}

\usepackage[T1]{fontenc}
\usepackage{graphicx}

\usepackage[utf8]{inputenc}

\usepackage{microtype}

\usepackage{caption}
\usepackage{multirow}
\usepackage{subcaption}
\usepackage{booktabs}
\usepackage{tabularx}
\usepackage{inconsolata}
\newcommand{\word}[1]{\emph{#1}}
\newcommand{\scone}{ScoNe}
\newcommand{\nonegbaseline}{Ignore-Negation}
\newcommand{\csr}{NLG}
\newcommand{\maf}{\texttt{MAF}-NLI}
\newcommand{\premvar}{\textbf{Premise}}
\newcommand{\hypvar}{\textbf{Hypothesis}}
\newcolumntype{P}[1]{>{\raggedright\arraybackslash}p{#1}}
\usepackage[]{titlesec}
\titlespacing*{\paragraph}{0ex}{1ex}{1ex}

\newcommand{\promptExample}[1]{
    {\centering
    \textbf{Prompt example}\\
    \framebox{
      \begin{minipage}[c]{0.95\textwidth}
        #1
    \end{minipage}}}}

\newcommand{\mydblnewline}{{\color{gray}{\textbackslash n\textbackslash n\\}}}
\newcommand{\mynewline}{{\color{gray}{\textbackslash n\\}}}

\newcommand{\uline}[1]{\underline{#1}}

\title{ScoNe: Benchmarking Negation Reasoning in Language Models\\ With Fine-Tuning and In-Context Learning%
\thanks{~~ \url{https://github.com/selenashe/ScoNe}}}

\author{Jingyuan Selena She \\ Haverford College\\  jshe@haverford.edu \\
\And
Christopher Potts\\ 
Stanford University\\  cgpotts@stanford.edu \\
\AND
Samuel R.~Bowman\\
New York University \& Anthropic, PBC\\
bowman@nyu.edu\\
\And
Atticus Geiger\\
Stanford University\\
atticusg@stanford.edu\\
}

\begin{document}
\maketitle

\begin{abstract}
A number of recent benchmarks seek to assess how well models handle natural language negation.
However, these benchmarks lack the controlled example paradigms that would allow us to infer whether a model had learned how negation morphemes semantically scope. 
To fill these analytical gaps, we present the \textbf{Sc}oped \textbf{Ne}gation NLI (\scone-NLI) benchmark, which contains contrast sets of six examples with up to two negations where either zero, one, or both negative morphemes affect the NLI label. 
We use \scone-NLI to assess fine-tuning and in-context learning strategies.
We find that RoBERTa and DeBERTa models solve \scone-NLI after many shot fine-tuning. For in-context learning, we test 
InstructGPT models and find that most prompt strategies are not successful, including those using step-by-step reasoning. To better understand this result, we extend \scone\ with \scone-\csr, a sentence completion test set that embeds negation reasoning in short narratives. Here, InstructGPT is successful, which reveals the model can correctly reason about negation, but struggles to do so on prompt-adapted NLI examples outside of its core pretraining regime.

\end{abstract}

\section{Introduction}
Negation is a ubiquitous but complex linguistic phenomenon that poses a significant challenge for NLP systems. A diverse array of benchmarks focused on negation have appeared in recent years, many of which contain families of contrasting examples that provide a local view of the model decision boundary \cite{Gardner:2020}. For instance, \citet{cooper1996using}, \citet{McCoy:2018}, \citet{Wang2019}, \citet{Ettinger2020}, \citet{hartmann-etal-2021-multilingual}, and \citet{kassner-schutze-2020-negated} all conduct evaluations with minimal pairs of examples that are identical except for  a negative morpheme. These examples reveal whether the presence of negation has a causal impact on model predictions. 

However, negation is not simply present or absent in a sentence. Rather, negation morphemes are semantic operators that take scope in complex ways, as we see in clear contrasts like \textit{the person who was at the talk wasn't happy} and \word{the person who wasn't at the talk was happy}. The recent CondaQA benchmark of \citet{Ravichander2022} includes minimal pairs aimed at determining whether models are sensitive to these differences in scope.

With the current paper, we seek to provide an even fuller picture of the complexities of negation and semantic scope. We introduce the English-language \textbf{Sco}ped \textbf{Ne}gation Natural Language Inference Benchmark (\scone-NLI). \scone-NLI extends the negated portion of the Monotonicity NLI dataset \cite{geiger-etal-2020-neural} such that each of the 1,202 examples is now a contrast set with six examples in which zero, one, or two negations are present and each negation may or may not have a semantic scope such that the NLI label is impacted by its presence. These six conditions offer a rich picture of how negation affects NLI reasoning, and they allow us to determine whether models are truly able to handle nested negation and scope or whether they have found simplistic solutions.

We evaluate models on \scone-NLI using many-shot fine-tuning as well as a wide range of in-context learning strategies. For fine-tuning approaches, we find that RoBERTa and DeBERTa models both solve \scone-NLI. For in-context learning, we evaluate the latest InstructGPT model with a variety of prompt strategies. We find that these models perform well on sections of \scone-NLI where the negation morphemes can simply be ignored, but they systematically fail in conditions where exactly one negative morpheme has semantic scope such that its presence changes the NLI label. In other words, these models fail to learn in context how negation actually takes scope.

\begin{table*}[t]
  \centering
\footnotesize
  \setlength{\tabcolsep}{3pt}
  \begin{subtable}{0.76\textwidth}
  \renewcommand{\arraystretch}{1.55}
  \begin{tabular}[c]{P{0.14\textwidth} P{0.32\textwidth} c P{0.3\textwidth} c}
    \toprule
    Split & Premise & Rel. & Hypothesis & Examples \\    
    \midrule
    No negation & The cowboy fell off a horse at the competition & $\sqsupset$ & The cowboy fell off a racehorse at the competition & 1,202 \\
    One Not Scoped & The cowboy did not fear anything, until he fell off a horse at the competition & $\sqsupset$ & The cowboy did not fear anything, until he fell off a racehorse at the competition& 1,202\\
    Two Not Scoped & The cowboy, who was not very old, was not proud that he fell off a horse at the competition & $\sqsupset$ & The cowboy, who was not very old, was not proud that he fell off a racehorse at the competition    & 1,202 \\
    Two Scoped & There is no way that the cowboy did not fall off a horse at the competition & $\sqsupset$ & There is no way that the cowboy did not fall off a racehorse at the competition & 1,202\\
    One Scoped & The cowboy did not fall off a horse at the competition & $\sqsubset$ & The cowboy did not fall off a racehorse at the competition  & 1,202\\
    One Scoped, One not Scoped & The cowboy did not fall off a horse, but the competition was not too important & $\sqsubset$ & The cowboy did not fall off a racehorse, but the competition was not too important& 1,202\\

    \bottomrule    
  \end{tabular}
  \caption{A six-example contrast set from \scone-NLI.}
  \label{tab:data-nli}
  \end{subtable}
  \begin{subtable}{0.23\textwidth}
    \centering
  \small
  \setlength{\tabcolsep}{4pt}
  \renewcommand{\arraystretch}{1.5}
  \begin{tabular}[c]{P{0.92\textwidth}}
  \toprule
\textbf{No Negation}\\
Glen is a fan of learning math. When he sees that his new high school requires that he take a calculus course, he\\
\textbf{Negation}\\
   Glen is not a fan of learning math. When he sees that his new high school requires that he take a calculus course, he\\
\textbf{Non-Scoping Negation}\\
Glen isn’t just a fan of learning math, he’s obsessive. When he sees that his new high school requires that he take a calculus course, he\\
\bottomrule
  \end{tabular}
  \caption{A three-example contrast set from \scone-\csr.}
  \label{tab:data-csr}
  \end{subtable}
  \caption{Two contrast sets from the \scone\ Benchmark}
  \label{tab:data}
\end{table*}

To better understand this result, we introduce a sentence completion test set (\scone-\csr) containing examples that seem better aligned with what we can infer about the training data used for InstructGPT models.
In each \scone-\csr\ example, negation reasoning is needed to provide a coherent ending to an incomplete narrative (see Figure~\ref{tab:data-csr}). \scone-\csr\ contains minimal triplets of examples where negation is absent, present with relevant scope, or present without relevant scope. InstructGPT is successful on \scone-\csr. When considered alongside our negative result for \scone-NLI, this finding seems to show that these models \textit{can} learn in-context about how negation takes scope, but only when the examples are hand-tailored to be aligned with the training data and aligned with known strengths of these models. Thus, when used together, \scone-NLI and \scone-\csr\ serve as a clear diagnostic for exploring useful prompting strategies and assessing the capacity of language models to reason about negation and scope.

\section{A Brief Review of Negation in NLI Benchmarks} A diverse array of benchmarks and diagnostic experiments have included negation reasoning in recent years \cite{nairn-etal-2006-computing, McCoy:2018, Wang2019, Ettinger2020, hartmann-etal-2021-multilingual, kassner-schutze-2020-negated, Ravichander2022}.

\citet{hossain-etal-2022-analysis} analyze a variety of natural language understanding benchmarks and find that negation is underrepresented, and that when negation is present it often has no impact on the example label. \citet{hossain-etal-2020-analysis} address this issue by manually adding negation to the premise-hypothesis pairs in MNLI \cite{williams-etal-2018-broad}, SNLI \cite{bowman-etal-2015-large}, and RTE \cite{Dagan2007ThePR, cooper1996using}.

\citet{yanaka-etal-2019-neural} introduce the crowd-sourced MED dataset, which has many NLI examples where negation generates inferences. Monotonicity NLI (MoNLI; \citealt{geiger-etal-2020-neural}) consists of modified SNLI sentences that have gold labels impacted by lexical entailments in affirmative contexts (PMoNLI) and lexical entailments reversed by a negation (NMoNLI).  BERT fine-tuned on SNLI and MNLI fails to generalize to both of these datasets, but succeeds with further fine-tuning on MED/MoNLI. Some automatically generated NLI datasets also include negation reasoning \cite{Geiger-etal:2019, richardson:2020, yanaka-etal-2019-help, yanaka-etal-2021-sygns}.

\begin{table*}
\centering
\footnotesize
\resizebox{\textwidth}{!}{
  \begin{tabular}[c]{c c c c c c c}
  \toprule
      & No &     One &    Two & Two & One & One Scoped, \\
     Fine-tuning Datasets
     & Negation & Not Scoped & Not Scoped & Scoped & Scoped & One not Scoped \\ 
  \midrule
   \maf\ & 82.0 & 86.0 & 81.5 & 91.0 & 5.0 & 5.0 \\
   \maf + MoNLI \cite{geiger-etal-2020-neural} & 96.2 & 87.5 & 99.5 & 8.9 & 100.0 & 100.0 \\
  \maf +  MED \cite{Yanaka2020DoNM} & 84.8 & 83.5 & 82.0 & 58.9 & 99.5 & 97.0 \\
  \maf + Neg-NLI \cite{hossain-etal-2020-analysis} & 91.3 & 88.5 & 83.0 & 70.4 & 37.0 & 29.0 \\
  \maf + MoNLI + \scone-NLI & 100.0 & 100.0 & 100.0 & 100.0 & 100.0 & 100.0 \\
  \bottomrule
  \end{tabular}
  }
  \caption{DeBERTa fine-tuning results on \scone-NLI. \maf\ stands for on MNLI, ANLI, and Fever-NLI.}
  \label{tab:DebertaResults}
\end{table*}
\begin{table}[t]
\footnotesize
  \centering
  \setlength{\tabcolsep}{3pt}
  \renewcommand{\arraystretch}{1.5}
  \begin{tabular}[c]{@{} P{0.12\textwidth} P{0.33\textwidth} @{}}
    \toprule
    Conditional Q & Is it true that if \premvar, then \hypvar? \\\midrule
    Hypothesis Q &  Assume that \premvar. Is it then definitely true that \hypvar? Answer yes or no. \\\midrule
    Conditional Truth & If \premvar, then \hypvar. Is this true? \\\midrule
    Brown et al. & P: \premvar{\textbackslash n}
Q: \hypvar {\textbackslash n} Yes, No, or Maybe?\\
    Structured & P: \premvar{\textbackslash n}
H: \hypvar{\textbackslash n}L:\\\midrule
 \multicolumn{2}{@{} l @{}}{Reasoning}   \\
\multicolumn{2}{@{} l @{}}{\parbox[t]{0.46\textwidth}{Logical and commonsense reasoning exam.{\textbackslash n}{\textbackslash n}\\
Explain your reasoning in detail, then answer with Yes or No. Your answers should follow this 4-line format:\mydblnewline
Premise: <a tricky logical statement about the world>.\mynewline
Question: <question requiring logical deduction>.\mynewline
Reasoning: <an explanation of what you understand about the possible scenarios>\mynewline
Answer: <Yes or No>.\mydblnewline
Premise: \premvar\mynewline
Question: \hypvar\mynewline
Reasoning: Let's think logically step by step. The premise basically tells us that}}\\
    \bottomrule    
  \end{tabular}
  \caption{Prompts used to adapt a 2-way NLI example (\premvar, \hypvar).
  Newlines are indicated with \textbackslash n. Full prompts with few-shot variants are in Appendix~\ref{app:prompts}.}
  \label{tab:prompt}
\end{table}

\section{\scone-NLI}

\scone-NLI is an extension of MoNLI \cite{geiger-etal-2020-neural}. MoNLI was generated by randomly selecting a sentence from SNLI and replacing a noun with a hypernym (more general term) or hyponym (less general term). The original and edited sentences are then used to form two premise--hypothesis pairs, one with the label \textit{entailment} and the other with the label \textit{neutral}. In about half of the examples, this replacement is in an affirmative context with no negation (PMoNLI). In the other half, it is under the scope of a single negation (NMoNLI). 

The authors generated \scone-NLI by using each example of NMoNLI to create a contrast set of six examples where gold labels are impacted by the scope of zero, one, or two negations, as in Table~\ref{tab:data}.
    
To succeed across all sections of \scone, models need to attend to the presence of negation as well as the way it scopes semantically. Table~\ref{tab:data-nli} shows an actual example of how \scone\ extends MoNLI. We use the train--test split of MoNLI where substituted lexical items are disjoint across training and testing data. Appendix~\ref{app:data} provides further details.

\begin{table*}[tp]
\small
\centering
\setlength{\tabcolsep}{7pt}
\begin{tabular}{@{} c l@{ \ } *{7}{c} @{}}
\toprule
&& No       & One  & Two        & Two    & One    & One Scoped,    &  \\
&& Negation & Not Scoped  & Not scoped & Scoped & Scoped & One not Scoped &   Overall \\
\midrule
\multirow{7}{*}{Zero-shot}
&Structured        &    0.50 &                 0.50 &                  0.50 &              0.50 &             0.50 &                                \uline{0.50} &  0.50 \\
&Brown et al.      &    0.74 &                 0.70 &                  0.74 &              0.55 &             0.44 &                                0.45 &  0.60 \\
&Conditional Q     &    0.79 &                 0.84 &                  0.80 &              0.50 &             0.52 &                                0.44 &  0.65 \\
&Conditional Truth &    \uline{0.98} &         0.86 &                  0.80 &              0.43 &             \uline{0.66} &                        0.47 &  0.70 \\
&Hypothesis Q      &    0.69 &                 \uline{0.90} &          0.70 &              0.51 &             0.62 &                                0.42 &  0.64 \\
&Reasoning         &    0.90 &                 0.88 &                  \uline{0.94} &      \uline{0.72} &     0.52 &                                0.46 &  \uline{0.73}
\\\midrule
\multirow{7}{*}{Few-shot}
&Structured        &    0.50 &                 0.50 &                  0.50 &              0.50 &             0.50 &                                \textbf{0.50} &  0.50 \\
&Brown et al.      &    0.86 &                 0.66 &                  0.80 &              0.83 &             0.36 &                                0.28 &  0.63 \\
&Conditional Q     &    0.92 &                 0.85 &                  0.90 &              0.62 &             0.34 &                                0.34 &  0.66 \\
&Conditional Truth &    0.94 &                 0.90 &                  0.94 &              0.64 &             0.36 &                                0.37 &  0.69 \\
&Hypothesis Q      &    0.98 &                 0.96 &                  0.94 &              0.83 &             0.51 &                                0.40 &  0.77 \\
&Reasoning         &    \textbf{0.99} &         \textbf{0.97} &          \textbf{0.98} &      \textbf{0.89} &     \textbf{0.69} &                        0.43 &  \textbf{0.82} \\
\midrule
& \nonegbaseline\ & 1.00 & 1.00 & 1.00 & 1.00 & 0.00 & 0.00 & 0.66\\
\bottomrule
\end{tabular}
\caption{In-context learning results on \scone-NLI for InstructGPT (\textit{davinci-003} engine; see Appendix~\ref{app:davinci002} for corresponding results for \textit{davinci-002}, which are uniformly lower). Zero-shot results are given in the first group of rows, with the best results in that condition underlined. Few-shot results are given in the second group, with the best results for this condition (and overall) in bold. The bottom row specifies a simple, idealized \nonegbaseline\ baseline  that makes predictions as if negations were absent. The baseline shows that the seemingly solid Overall results of these models are driven largely by conditions for which negation can be ignored. Conversely, models are often at or below chance where negation is critical in some way.}
\label{tab:nlifew}
\end{table*}
\begin{table}
\setlength{\tabcolsep}{7pt}
\small
\centering
\begin{tabular}{@{} l c c c c @{}}
\toprule
& No        & One    & One Not & \\
& Negation  & Scoped & Scoped  & Overall \\
\midrule
Zero-shot &  0.99 &  0.90 & 0.88 &  0.92 \\
Few-shot  &  0.93 &  1.00 & 0.93 &  0.95 \\
\bottomrule
\end{tabular}
\caption{Results for \scone-\csr\ using \texttt{davinci-003}. 
The three conditions correspond to those of \scone\ and test the essential scope-taking properties of negation. 
}
\label{tab:scone-csr-results}
\end{table}

\paragraph{Fine-Tuning on \scone-NLI}
We used publicly available weights on HuggingFace for the DeBERTa-v3-base models already fine-tuned on MNLI, Fever-NLI, and Adversarial-NLI \cite{laurer2022,he2021deberta}. Appendix~\ref{app:roberta} contains comparable results for the RoBERTa model \cite{liu2019roberta}. Fine-tuning results are in Table~\ref{tab:DebertaResults}.

Fine-tuning on existing NLI datasets is insufficient for good performance on \scone-NLI:
DeBERTa-v3-base fine-tuned on existing NLI datasets, even those that focus on negation, systematically fails. Thus, it seems that \scone-NLI captures novel aspects of negation reasoning.

In contrast, fine-tuning on MoNLI and \scone-NLI training data results in near perfect performance on \scone-NLI test data. This shows that DeBERTa can learn negation reasoning and generalize to new lexical items.

\paragraph{In-context Learning on \scone-NLI}\label{sec:experiment}

We evaluated InstructGPT using OpenAI's API with \word{text-davinci-002} and \word{text-davinci-003} engines and a temperature of 0.0 \cite{brown2020}. We ask InstructGPT to infer NLI labels given the premise and hypothesis using prompts. All prompts are constructed such that if the response contain ``yes'' (case-insensitive), then the label \word{entailment} is predicted, else the label \word{neutral} is predicted. We use six prompts (Table~\ref{tab:prompt}). For each prompt, we implemented both zero-shot and few-shot inference experiments. Appendix~\ref{app:prompts} provides the full prompts.

\paragraph{InstructGPT makes systematic errors similar to a baseline that ignores negation entirely.} The best results are for the few-shot reasoning prompt with \word{davinci-003}. While its overall accuracy of 82\% may initially appear to be a success, further analysis reveals otherwise. InstructGPT succeeds only on the sections of \scone-NLI where zero or two negations take scope, namely, no negation (99\%), one not scoped (97\%), two not scoped (98\%), and two scoped (89\%). InstructGPT performs much worse on sections where exactly one negation takes scope, namely one scoped (69\%), one scoped/one not (48\%). An idealized baseline entirely ignoring the presence of negation (last row of Table~\ref{tab:nlifew}) succeeds and fails on the same sections, indicating a systematic flaw in InstructGPT. %

\section{\scone-\csr\ }\label{sec:scone-nlg}

InstructGPT fails to reason about negation when given NLI examples that must be adapted to natural language generation (NLG) with prompts. We hypothesized that InstructGPT may correctly reason about negation when evaluated on examples hand tailored to its pretraining objective, because there is no need for prompt engineering \cite{Liu2021PretrainPA, wei2022chain, Kojima2022LargeLM}.

\paragraph{Dataset}
ScoNe-\csr\ is a natural language generation dataset that contains 74 contrasting triplets of examples of half-completed naturalistic narratives that have different coherent completions depending on the presence and scope of a negation. InstructGPT fails on the sections of \scone-NLI examples containing only one negation,  so we opt for contrast sets with three examples that require knowledge of a lexical entailment in an affirmative context without negation, an affirmative context with non-scoping negation, and an negative context with scoping negation, respectively. See Table~\ref{tab:data-csr}. 

\paragraph{In-context Learning on \scone-NLG}
We used InstructGPT to complete the partial sentence inputs with the \word{text-davinci-003} engine (temperature of 0.0). In the zero-shot setting, the prompt consists of the \scone-\csr\ example. In the few-shot setting, four demonstrations from \scone-\csr\ are given one with no negation, two with scoping negation, and one with non-scoping negation. See Appendix~\ref{app:scone-nlg-prompts} for the complete prompts.

To evaluate, the authors went through the responses by hand and determined whether the generated text is coherent and compatible with the initial narrative. The authors agreed on these annotations for 216/222 of the zero-shot responses with a Fleiss kappa of 0.84 and 220/222 of the few-shot responses with a Fleiss kappa of 0.91. These agreement rates are so high that we evaluate InstructGPT only for the cases where the annotators agree.  Here, InstructGPT is successful but not perfect, achieving 95\% and 92\% accuracy in the few and zero-shot settings, respectively. We do not observe the systematic failures seen on \scone-NLI.

\begin{figure*}
\begin{subfigure}[b]{0.23\textwidth}    
\scriptsize
\begin{codebox}
\Procname{$\proc{ScoNe-Bool}(\textbf{p}, \textbf{h})$}
\li $\textit{lexrel} \leftarrow \proc{get-lexrel}(\textbf{p}, \textbf{h})$
\li $\textit{neg1} \leftarrow \proc{first-scope}(\textbf{p}, \textbf{h})$
\li $\textit{neg2} \leftarrow \proc{second-scope}(\textbf{p}, \textbf{h})$
\li $\textbf{if} \; (\textit{neg1} \; \oplus \; \textit{neg2}))$:
\li \; \; \Return $\proc{reverse}(\textit{lexrel} )$ \End
\li \Return $\textit{lexrel}$
\end{codebox}
\caption{An interpretable program that solves \scone-NLI by computing two Boolean variables that encode whether the first and second negation scope and reversing entailment if exactly one is true.}
\label{fig:algsolbool}
\end{subfigure}
\hfill
\begin{subfigure}[b]{0.23\textwidth}    
\scriptsize
\begin{codebox}
\Procname{$\proc{ScoNe-Count}(\textbf{p}, \textbf{h})$}
\li $\textit{lexrel} \leftarrow \proc{get-lexrel}(\textbf{p}, \textbf{h})$
\li $\textit{count} \leftarrow \proc{count-scoped}(\textbf{p}, \textbf{h})$
\li $\textbf{if}$ \textit{count} == 1:
\li \; \; \Return $\proc{reverse}(\textit{lexrel} )$ \End
\li \Return $\textit{lexrel}$
\end{codebox}
\caption{An interpretable program that solves \scone-NLI by counting the scoped negations and reversing entailment if there is exactly one.}
\label{fig:algsolcount}
\end{subfigure}
\hfill
\begin{subfigure}[b]{0.23\textwidth}
\scriptsize
\begin{codebox}

\Procname{$\proc{Ignore-Scope}(\textbf{p}, \textbf{h})$}
\li $\textit{lexrel} \leftarrow \proc{get-lexrel}(\textbf{p}, \textbf{h})$
\li $\textit{count} \leftarrow \proc{count-neg}(\textbf{p}, \textbf{h})$
\li $\textbf{if}$ \textit{count} == 1:
\li \; \; \Return $\proc{reverse}(\textit{lexrel} )$ \End
\li \Return $\textit{lexrel}$
\end{codebox}
\caption{A flawed heuristic program: we count the negations and reverse entailment if there is a single negation, which is equivalent to ignoring the scope of negation.}
\label{fig:algnoscope}
\end{subfigure}
\hfill
\begin{subfigure}[b]{0.23\textwidth}   
\scriptsize
\begin{codebox}
\Procname{$\proc{Ignore-Negation}(\textbf{p}, \textbf{h})$}
\li $\textit{lexrel} \leftarrow \proc{get-lexrel}(\textbf{p}, \textbf{h})$
\li \Return $\textit{lexrel}$
\end{codebox}
\caption{A flawed heuristic program for \scone-NLI that outputs the lexical relation and ignores negation entirely.}
\label{fig:algnonegation}
\end{subfigure}
\caption{Four human-interpretable algorithms for \scone-NLI. The first two solve the task perfectly, and the other two implement flawed heuristics that a model might learn to implement. The function \textsc{get-lexrel} retrieves the relation between the aligned words in the premise and hypothesis, \textsc{count-scoped} counts scoped negations, \textsc{count-neg} counts negations regardless of scope, and \textsc{get-first} returns true if the first negation scopes, while \textsc{get-second} returns true if there is a second negation and it scopes.}
\label{fig:algs}
\end{figure*}
\section{Future Work on Interpretability }\label{sec:interpret}

\scone\ is based in naturalistic examples, but it also has a controlled structure that offers valuable opportunities to move beyond simple behavioral testing and more deeply understand \textit{how} models solve tasks related to lexical entailment and negation. 

The theory of causal abstraction provides a framework for interpretability \cite{geigericard}, where a neural model can be understood to implement the intermediate variables and internal structure of a program or algorithm \cite{Geiger:2021, geiger-etal-2021-iit, wu-etal-2022-causal, wu-etal-2022-cpm, Huang-etal:2022, Geiger-etal:2023:DAS}. In fact, the MoNLI dataset and the technique of interchange interventions (which is the primary technique in causal abstraction analysis) were jointly introduced in \citealt{geiger-etal-2020-neural}, where interchange interventions were used to investigate whether a BERT model implements a simple, human-interpretable algorithm that can perfectly label MoNLI using a variable representing lexical entailment and a variable representing the presence of negation. 

With \scone, we can ask even deeper interpretability questions of this form. To encourage future work in this direction, we present a range of algorithmic solutions in Figure~\ref{fig:algs}. Two of these solutions solve \scone\ and could perhaps explain neural models that learn the task perfectly, and two others implement flawed heuristics that could explain neural models with poor task performance.

Figure~\ref{fig:algsolbool} and Figure~\ref{fig:algsolcount} present two intuitive and correct algorithms that solve \scone, but have distinct intermediate variables and internal structure. The first computes two Booleans representing whether each negation scopes, and the second computes a count of how many negations scope.

Figure~\ref{fig:algnonegation} is the flawed heuristic that ignores negation that we discussed in Section~\ref{sec:experiment} as a hypothesis about how models fail at our task. Figure~\ref{fig:algnonegation} is a second flawed heuristic that counts the number of negations present but ignores scope.
 
Using the toolkit of causal abstraction, we can assess models not only behaviorally, but also evaluate whether they implement an interpretable algorithm. The results of \citet{Geiger-etal:2023:DAS} begin to show how such analyses could be extended to in-context learning with LLMs, as in Section~\ref{sec:scone-nlg}.

\section{Conclusion}
We introduced \scone, a benchmark for fine-tuning and in-context learning experiments on negation. \scone\ is challenging for NLI models fine-tuned on other datasets, even those designed for negation reasoning, but modest amount of fine-tuning on \scone\ leads to success. For in-context learning, we find that that InstructGPT models fail dramatically on \scone. However, we also introduce \scone-\csr, which uses more narrative-like examples to probe models' capacity to handle negation, and show that InstructGPT is successful with zero-shot and few-shot prompts for this task. These results show that \scone\ supports fine-grained assessments of whether models can reason accurately about natural language negation, and our discussion in Section~\ref{sec:interpret} suggests that \scone\ can be a powerful tool for discovering \emph{how} models reason semantically.

\section*{Limitations}

We are releasing \scone\ as a diagnostic tool for conducting controlled scientific experiments. This is our primary intended use, and we advise against uncritical use of \scone\ for real-world applications, as we have not audited the dataset for such purposes.

As a diagnostic tool, \scone's primary limitation is its focus on English. Cross-linguistically, we find many strategies for expressing negation. The English-language strategy of using mostly adverbial modifiers for sentential negation is not the only one by any means, and we would expect to see quite different results for languages in which negation is expressed, for example, with verbal suffixes. This highlights the value of potential future efforts extending \scone\ to other languages.

By the same token, we acknowledge that  many linguistic phenomena interact with negation even internal to English. \scone\ restricts to negation in the context of lexical entailment, and mostly uses ``not'' as the negative morpheme. This excludes a wide range of negation morphemes and negation strategies that ultimately need to be brought into the picture.

Finally, we note that there may be undesirable biases in \scone\ that could interact with biases in the models. \scone\ is in part derived from SNLI, which is known to contain gaps, social biases, and artifacts \cite{poliak2018, mccoy2019, belinkov2019, gururangan2018, tsuchiya2018}, and \scone\ may inherit some of these.

\bibliographystyle{acl_natbib}
\bibliography{anthology,custom}

\newpage
\clearpage
\onecolumn
\appendix

\section*{Appendices}

\section{Experimental Details}
\subsection{Fine-tuning Protocol} For our fine-tuning experiments, we used a learning rate of 1e-5, batch size of 4, gradient accumulation steps of 6 for a total of 10 epochs. We used these default hyperparameters as they were successful in fine-tuning on ScoNe. We implemented these experiments with Pytorch \cite{pytorch} and used the scikit learn package \cite{scikit-learn}. 
 
 \subsection{Hugging Face Models} We test RoBERTa\footnote{released under the MIT license} and DeBERTa\footnote{released under the MIT license} in these experiments. We used the roberta-large model fine-tuned on MNLI\footnote{released under the MIT license} with 354 million parameters, 500K steps, and trained on 1,024 V100 GPUs \cite{liu2019roberta}. DeBERTa-v3-base-mnli-fever-anli model\footnote{released under the MIT license} was fine-tuned on MNLI, Fever-NLI,\footnote{released under the Creative Commons Attribution-ShareAlike License (version 3.0)} and ANLI.\footnote{released under the Attribution-NonCommercial 4.0 International license} 
 \newline

RoBERTa weights link:

\url{https://huggingface.co/roberta-large-mnli}
 \newline

Deberta weights link: 

\url{https://huggingface.co/MoritzLaurer/DeBERTa-v3-base-mnli-fever-anli}
\newline

\subsection{Fine-Tuning Datasets}

We further fine-tuned our model on the datasets MoNLI,\footnote{released under the Creative Commons Attribution Share Alike 4.0 International license} Negation-NLI, \footnote{released under the MIT license} MED. \footnote{released under the Creative Commons Attribution Share Alike 4.0 International license}

\section{RoBERTa Results}\label{app:roberta}

\begin{table*}[h]
\centering
\footnotesize
  \begin{tabular}[c]{c c c c c c c}
  \toprule
      & No & One & Two & Two & One & One Scoped, \\ 
     Fine-tuning Datasets & Negation &Not Scoped & Not Scoped & Scoped & Scoped & One not Scoped\\ 
  \midrule
   \maf\ & 96.5 & 97.0 & 97.0 & 96.5 & 3.0 & 5.0\\
   \maf + MoNLI \cite{geiger-etal-2020-neural} & 85.4 & 100.0 & 100.0 & 4.5 & 100.0 & 100.0\\
  \maf +  MED \cite{Yanaka2020DoNM} & 85.1 & 92.0 & 89.5 & 44.6 & 85.5 & 81.5\\
  \maf + Neg-NLI \cite{hossain-etal-2020-analysis} & 93.1 & 97.5 & 93.0 & 73.2 & 20.5 & 17.5\\
  \maf + MoNLI + \scone-NLI & 100.0 & 100.0 & 100.0 & 100.0 & 100.0 & 100.0\\
  \bottomrule
  \end{tabular}
  \caption{RoBERTa fine-tuning results on \scone-NLI. \maf\ stands for on MNLI, ANLI, and Fever-NLI.}
  \label{tab:RobertaResults}
\end{table*}

\section{\scone\ Dataset Details}\label{app:data}
For some examples, we modified the lexical items replaced. Consider the NMoNLI sentence pair ‘a man is not tossing anything’-‘a man is not tossing socks’ (entailment), and non-scoping counterpart ‘a man not here is tossing something’-‘a man not here is tossing socks’ (neutral). Here, 'anything' must be replaced by 'something'. The positive and negative examples in MoNLI \textit{do not} come in minimal pairs, so the examples in \scone-NLI with no negation are \textit{not} from PMoNLI.

\section{Prompting Methods}\label{app-prompt-methods}

The experimental runs reported in the paper were conducted on January 11, 2023. We used InstructGPT\footnote{information on terms of use is available at: https://openai.com/terms/}  models with 1.3 billion parameters and 6 billion parameter. The exact cost of constructing the InstructGPT models is not public, but the pre-training protocol involves (1) fine-tuning a GPT3 model on an instruction following dataset, (2) fine-tuning a GPT3 model to rank different answers to the instruction following dataset,  and (3) using reenforcement learning to combine these two models. We use a temperature parameter of 0.0 for all experiments. If the response contains ``yes'' (case-insensitive), then we infer the label \texttt{entailment}, else we infer \texttt{neutral}. Across experiments, the only thing that varies is the nature of the prompt function.

\section{In-Context Learning Prompts}\label{app:prompts}

We have indicated all actual newlines with {\color{gray}\textbackslash n}. The newlines in the formatting are just to make them intuitive to read.

\subsection{Conditional Question Prompt}

\promptExample{Is it true that if we didn't eat pizza, then we didn't eat food?}

\subsection{Few-Shot Conditional Question Prompt}

\promptExample{Q1: Is it true that if a not so tall person reading a paper is not currently sitting inside a building, then a not so tall person reading a paper is not currently sitting inside a club?\mynewline
A1: Yes\mynewline
\mynewline
Q2: Is it true that if the man does not own a dog and does not own a cat, then the man does not own a retriever and does not own a cat?\mynewline
A2: Yes\mynewline
\mynewline
Q3: Is it true that if a not so tall person reading a paper is not currently sitting inside a cabin, then a not so tall person reading a paper is not currently sitting inside a building?\mynewline
A3: Maybe\mynewline
\mynewline
Q4: Is it true that if a not so tall person reading a paper is not currently sitting inside a casino, then a not so tall person reading a paper is not currently sitting inside a building?
A4: Maybe\mynewline
\mynewline
Q: Is it true that if we didn't eat pizza, then we didn't eat food?\mynewline
A:}

\subsection{Hypothesis Question Prompt}

\promptExample{Assume that we didn't eat pizza. Is it then definitely true that we didn't eat food? Answer Yes or No.}

\newpage
\subsection{Few-Shot Hypothesis Question Prompt}

\promptExample{Q1: Assume that a not so tall person reading a paper is not currently sitting inside a building. Is it then definitely true that a not so tall person reading a paper is not currently sitting inside a casino? Answer Yes or No.\mynewline
A1: Yes\mynewline
\mynewline
Q2: Assume that the girl will not get a stuffed dog as a gift, but not because she failed the exam. Is it then definitely true that the girl will not get a stuffed pinscher as a gift, but not because she failed the exam? Answer Yes or No.\mynewline
A2: Yes\mynewline
\mynewline
Q3: Assume that the girl will not get a stuffed shetland as a gift, but not because she failed the exam. Is it then definitely true that the girl will not get a stuffed dog as a gift, but not because she failed the exam? Answer Yes or No.\mynewline
A3: No\mynewline
\mynewline
Q4: Assume that a not so tall person reading a paper is not currently sitting inside a monastery. Is it then definitely true that a not so tall person reading a paper is not currently sitting inside a building? Answer Yes or No.\mynewline
A4: No\mynewline
\mynewline
Q: Assume that we didn't eat pizza. Is it then definitely true that we didn't eat food? Answer Yes or No.\mynewline
A:}
\subsection{Conditional Truth Evaluation Prompt}

\promptExample{If we didn't eat pizza, then we didn't eat food. Is this true?}
\newpage

\subsection{Few-Shot Conditional Truth Evaluation Prompt}

\promptExample{C1: If the man does not own a dog and does not own a cat, then the man does not own a shetland and does not own a cat. Is this true?\mynewline
A1: Yes\mynewline
\mynewline
C2: If a not so tall person reading a paper is not currently sitting inside a building, then a not so tall person reading a paper is not currently sitting inside a house. Is this true?\mynewline
A2: Yes\mynewline
\mynewline
C3: If the man does not own a collie and does not own a cat, then the man does not own a dog and does not own a cat. Is this true?\mynewline
A3: Maybe\mynewline
\mynewline
C4: If the man does not own a corgi and does not own a cat, then the man does not own a dog and does not own a cat. Is this true?\mynewline
A4: Maybe\mynewline
\mynewline
C:If we didn't eat pizza, then we didn't eat food. Is this true?\mynewline
A:}

\subsection{Brown Et Al Style Prompt}

\promptExample{C: We didn't eat pizza\mynewline
Q: We didn't eat food. Yes, No, or Maybe?}

\subsection{Few-Shot Brown Et Al Style Prompt}

\promptExample{C1: The man, who's eyes are not open, is not steering a car.\mynewline
Q1: The man, who's eyes are not open, is not steering a sedan. Yes, No, or Maybe?\mynewline
A2: Yes\mynewline
\mynewline
C2: A dog not on the playground did not catch any ball.\mynewline
Q2: A dog not on the playground did not catch any volleyball. Yes, No, or Maybe?\mynewline
A3: Yes\mynewline
\mynewline
C3: the man does not own a collie and does not own a cat.\mynewline
Q3: the man does not own a dog and does not own a cat. Yes, No, or Maybe?\mynewline
A4: Maybe\mynewline
\mynewline
C4: A not so tall person reading a paper is not currently sitting inside a inn.\mynewline
Q4: A not so tall person reading a paper is not currently sitting inside a building. Yes, No, or Maybe?\mynewline
A5: Maybe\mynewline
\mynewline
C: We didn't eat pizza\mynewline
Q: We didn't eat food. Yes, No, or Maybe?\mynewline
A:}
\newpage

\subsection{Structured Prompt}

\promptExample{P: We didn't eat pizza\mynewline
H: We didn't eat food\mynewline
L:}

\subsection{Few-Shot Structured Prompt}

\promptExample{P1: The players who did not score did not have a ball.\mynewline
H1: The players who did not score did not have a baseball.\mynewline
L1: entailment\mynewline
\mynewline
P2: the man does not own a dog and does not own a cat.\mynewline
H2: the man does not own a poodle and does not own a cat.\mynewline
L2: entailment\mynewline
\mynewline
P3: the man does not own a terrier and does not own a cat.\mynewline
H3: the man does not own a dog and does not own a cat.\mynewline
L3: neutral\mynewline
\mynewline
P4: the man does not own a husky and does not own a cat.\mynewline
H4: the man does not own a dog and does not own a cat.\mynewline
L4: neutral\mynewline
\mynewline
P: We didn't eat pizza\mynewline
H: We didn't eat food\mynewline
L:}

\subsection{Reasoning Prompt}

\promptExample{Logical and commonsense reasoning exam.\mynewline
\mynewline
Explain your reasoning in detail, then answer with Yes or No. Your answers should follow this 4-line format:\mynewline
\mynewline
Premise: <a tricky logical statement about the world>.\mynewline
Question: <question requiring logical deduction>.\mynewline
Reasoning: <an explanation of what you understand about the possible scenarios>.\mynewline
Answer: <Yes or No>.\mynewline
\mynewline
Premise: we didn't eat pizza\mynewline
Question: Can we logically conclude for sure that we didn't eat food?\mynewline
Reasoning: Let's think logically step by step. The premise basically tells us that}
\newpage

\subsection{Few-shot Reasoning Prompt}

For this prompt, we insert two demonstrations right before the test example. These are of the correct type for the test example, and they exemplify each of the two labels. The demonstrations are from a fixed set of examples, which we include here:

\subsubsection{No Negation}
\promptExample{Here are some examples of the kind of reasoning you should do:\mynewline
\mynewline
Premise: The students ate pizza\mynewline
Question: Can we logically conclude for sure that the students ate food?\mynewline
Reasoning: Let's think logically step by step. The premise basically tells us that pizza is a type of food. Therefore, the premise that the students ate pizza entails that the students ate food.\mynewline
Answer: Yes\mynewline
\mynewline
Premise: The students ate food\mynewline
Question: Can we logically conclude for sure that the students ate pizza?\mynewline
Reasoning: Let's think logically step by step. The premise basically tells us that pizza is a type of food. Therefore, the premise that the students ate food does not allow us to conclude that the students ate pizza. They might have eaten something else.\mynewline
Answer: No\mynewline\mynewline}

\subsubsection{One Scoped}
\promptExample{Here are some examples of the kind of reasoning you should do:\mynewline
\mynewline
Premise: The students didn't eat any pizza\mynewline
Question: Can we logically conclude for sure that the students didn't eat any food?\mynewline
Reasoning: Let's think logically step by step. The premise basically tells us that pizza is a type of food. Therefore, the premise that the students didn't eat any pizza does not allow us to conclude that the students didn't eat any food. They might have eaten something else.\mynewline
Answer: No\mynewline
\mynewline
Premise: The students didn't eat any food\mynewline
Question: Can we logically conclude for sure that the students didn't eat any pizza?\mynewline
Reasoning: Let's think logically step by step. The premise basically tells us that pizza is a type of food. Therefore, the premise that the students didn't eat any food entails that the students didn't eat any pizza.\mynewline
Answer: Yes\mynewline\mynewline
}
\newpage

\subsubsection{One Not Scoped}
\promptExample{Here are some examples of the kind of reasoning you should do:\mynewline
\mynewline
Premise: The students who weren't in class ate pizza\mynewline
Question: Can we logically conclude for sure that the students who weren't in class ate food?\mynewline
Reasoning: Let's think logically step by step. The premise basically tells us that pizza is a type of food. Therefore, the premise that the students who weren't in class ate pizza entails that the students who weren't in class ate food.\mynewline
Answer: Yes\mynewline
\mynewline
Premise: The students who weren't in class ate food\mynewline
Question: Can we logically conclude for sure that the students who weren't in class ate pizza?\mynewline
Reasoning: Let's think logically step by step. The premise basically tells us that pizza is a type of food. Therefore, the premise that the students who weren't in class ate food does not allow us to conclude that the students who weren't in class ate pizza. They might have eaten something else.\mynewline
Answer: No\mynewline\mynewline
}

\subsubsection{One Scoped, One Not Scoped}
\promptExample{Here are some examples of the kind of reasoning you should do:\mynewline
\mynewline
Premise: The students who weren't in class didn't eat any pizza\mynewline
Question: Can we logically conclude for sure that the students who weren't in class didn't eat any food?\mynewline
Reasoning: Let's think logically step by step. The premise basically tells us that pizza is a type of food. Therefore, the premise that the students who weren't in class didn't eat any pizza does not allow us to conclude that the students who weren't in class didn't eat any food. They might have eaten something else.\mynewline
Answer: No\mynewline
\mynewline
Premise: The students who weren't in class didn't eat any food\mynewline
Question: Can we logically conclude for sure that the students who weren't in class didn't eat any pizza?\mynewline
Reasoning: Let's think logically step by step. The premise basically tells us that pizza is a type of food. Therefore, the premise that the students who weren't in class didn't eat any food entails that the students who weren't in class didn't eat any pizza.\mynewline
Answer: Yes\mynewline\mynewline
}

\newpage
\subsubsection{Two Not Scoped}
\promptExample{Here are some examples of the kind of reasoning you should do:\mynewline
\mynewline
Premise: The students who weren't in class ate pizza that wasn't hot\mynewline
Question: Can we logically conclude for sure that the students who weren't in class ate food that wasn't hot?\mynewline
Reasoning: Let's think logically step by step. The premise basically tells us that pizza is a type of food. Therefore, the premise that the students who weren't in class ate pizza that wasn't hot entails that the students who weren't in class ate food that wasn't hot.\mynewline
Answer: Yes\mynewline
\mynewline
Premise: The students who weren't in class ate food that wasn't hot\mynewline
Question: Can we logically conclude for sure that the students who weren't in class ate pizza that wasn't hot?\mynewline
Reasoning: Let's think logically step by step. The premise basically tells us that pizza is a type of food. Therefore, the premise that the students who weren't in class ate food that wasn't hot does not allow us to conclude that the students who weren't in class ate pizza that wasn't hot. They might have eaten something else.\mynewline
Answer: No\mynewline\mynewline
}

\subsubsection{Two Scoped}
\promptExample{Here are some examples of the kind of reasoning you should do:\mynewline
\mynewline
Premise: It is not the case that the students didn't eat any pizza\mynewline
Question: Can we logically conclude for sure that it is not the case that the students didn't eat any food?\mynewline
Reasoning: Let's think logically step by step. The premise basically tells us that pizza is a type of food. Therefore, the premise that it is not the case that the students didn't eat any pizza entails that it is not the case that the students didn't eat any food.\mynewline
Answer: Yes\mynewline
\mynewline
Premise: It is not the case that the students didn't eat any food\mynewline
Question: Can we logically conclude for sure that it is not the case that the students didn't eat any pizza?
Reasoning: Let's think logically step by step. The premise basically tells us that pizza is a type of food. Therefore, the premise that it is not the case that the students didn't eat any food does not allow us to conclude that it is not the case that the students didn't eat any pizza. They might have eaten something else.\mynewline
Answer: No\mynewline\mynewline
}

\newpage
\subsection{\scone-\csr\ Prompts}\label{app:scone-nlg-prompts}

In the zero-shot condition, models are simply prompted with the \scone-\csr\ examples. In the few-shot condition, the test is example is proceeded with a fixed set of four demonstrations, separated by double newlines. The examples are as follows:

\promptExample{
Glen is not a fan of learning math. When he sees that his new high school requires that he take a geometry course, he is not pleased.\mynewline
\mynewline
I saw John take his BMW to the store the other day, so when Suzy asked me if John owns a car, I said yes.\mynewline
\mynewline
I’ve seen John with a dog that isn’t very cute, so when Suzy asked me if John owns a pet, I said yes.\mynewline
\mynewline
I recently confirmed that John is not allergic to any shellfish. So it makes sense that when we served shrimp}

\section{In-Context Learning Results for davinci-002}\label{app:davinci002}

\begin{table*}[h]
\small
\centering
\setlength{\tabcolsep}{7pt}
\begin{tabular}{@{} c l@{ \ } *{7}{c} @{}}
\toprule
&& No       & One  & Two        & Two    & One    & One Scoped,    &  \\
&& Negation & Not Scoped  & Not scoped & Scoped & Scoped & One not Scoped &   Overall \\
\midrule
\multirow{7}{*}{Zero-shot}
&Structured        &    0.50 &                 0.50 &                  0.50 &              0.50 &             0.50 &                  0.50 &  0.50 \\
&Brown et al.      &    0.69 &                 0.60 &                  0.59 &              0.55 &             0.50 &                  0.48 &  0.57 \\
&Conditional Q     &    0.76 &                 0.55 &                  0.65 &              0.50 &             0.50 &                  0.50 &  0.58 \\
&Conditional Truth &    0.76 &                 0.64 &                  0.66 &              0.60 &             0.50 &                  \uline{0.57} &  0.62 \\
&Hypothesis Q      &    0.80 &                 \uline{0.83} &          \uline{0.86} &      \uline{0.62} &     0.45 &                  0.40 &  \uline{0.66} \\
&Reasoning         &    \uline{0.85} &         0.70 &                  0.68 &              \uline{0.62} &     \uline{0.57} &          0.56 &  \uline{0.66} \\
\midrule
\multirow{7}{*}{Few-shot}
&Structured        &    0.50 &                 0.50 &                  0.50 &              0.50 &             0.50 &                     0.50 &  0.50 \\
&Brown et al.      &    0.82 &                 0.75 &                  0.78 &              \textbf{0.72} &    0.35 &                     0.29 &  0.62 \\
&Conditional Q     &    0.92 &                 0.82 &                  0.78 &              0.52 &             0.36 &                     0.32 &  0.62 \\
&Conditional Truth &    0.92 &                 0.89 &                  0.88 &              0.59 &             0.36 &                     0.37 &  0.67 \\
&Hypothesis Q      &    \textbf{0.99} &        \textbf{0.91} &         \textbf{0.92} &     0.68 &             0.38 &                     0.40 &  \textbf{0.72} \\
&Reasoning         &    0.73 &                 0.85 &                  0.78 &              0.62 &             \textbf{0.74} &            \textbf{0.54} &  0.71 \\
\bottomrule
\end{tabular}
\caption{In-context learning results for GPT-3 (davinci-002 engine).}
\end{table*}

\end{document}